\documentclass[10pt,twocolumn,letterpaper]{article}

\usepackage[pagenumbers]{cvpr} % To force page numbers, e.g. for an arXiv version

\usepackage[dvipsnames]{xcolor}

\definecolor{cvprblue}{rgb}{0.21,0.49,0.74}
\usepackage[pagebackref,breaklinks,colorlinks,citecolor=cvprblue]{hyperref}
\usepackage{cuted}
\usepackage{multirow}
\usepackage[accsupp]{axessibility} % Improves PDF readability for those with visual impairments.

\def\paperID{3718} % *** Enter the Paper ID here
\def\confName{CVPR}
\def\confYear{2024}

\title{Texture-Preserving Diffusion Models for High-Fidelity Virtual Try-On}

\author{
  Xu Yang$^1$ \quad
  Changxing Ding$^{1*}$ \quad
  Zhibin Hong$^2$ \quad
  Junhao Huang$^2$ \quad
  Jin Tao$^1$ \quad
  Xiangmin Xu$^1$ \\
  $^1$South China University of Technology \quad
  $^2$S Research \\ 
  {\tt\small ftyang\_xu@mail.scut.edu.cn} \quad
  {\tt\small chxding@scut.edu.cn} \quad
  {\tt\small zhib.hong@gmail.com} \\
  {\tt\small junhao.huang.77@gmail.com} \quad
  {\tt\small arjtao@scut.edu.cn} \quad
  {\tt\small xmxu@scut.edu.cn} 
}

\begin{document}

\maketitle

\begin{strip}
    \vspace*{-15mm}
    \centering
    \includegraphics[width=1\textwidth]{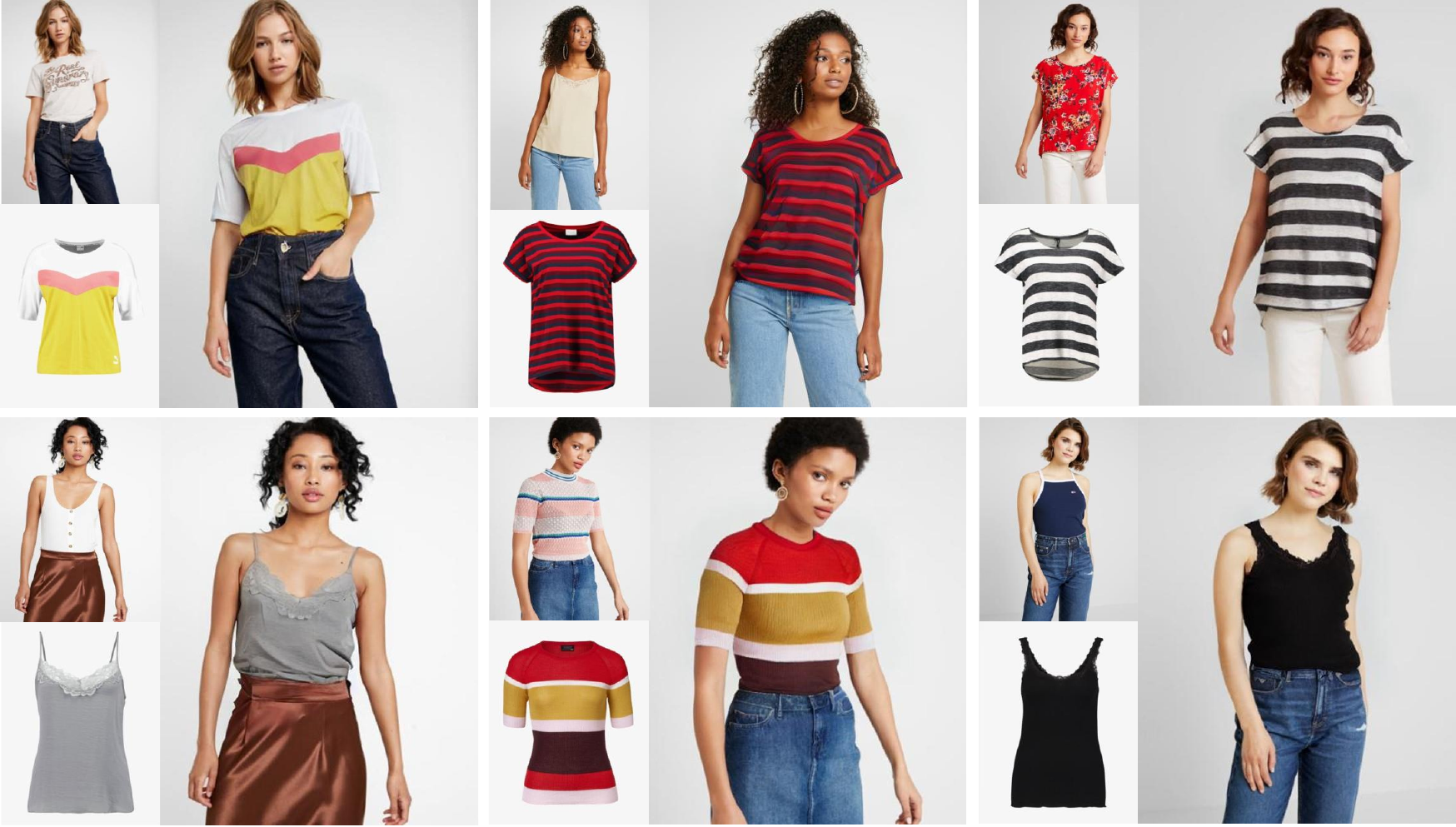}
    \captionof{figure}{The sample try-on images synthesized by our Texture-Preserving Diffusion (TPD) model. In each triplet, the two left images are the original person and garment images from VITON-HD~\cite{choi2021viton} database. The right one depicts the synthesized image.}
    \label{fig_introduction}
    \vspace*{-2mm}
\end{strip}

\makeatletter
\renewcommand*{\@makefnmark}{}
\footnotetext{$^*$Corresponding author.}
\renewcommand*{\@makefnmark}{\hss\@textsuperscript{\normalfont\@thefnmark}}
\makeatother

\begin{abstract}
Image-based virtual try-on is an increasingly important task for online shopping. It aims to synthesize images of a specific person wearing a specified garment. Diffusion model-based approaches have recently become popular, as they are excellent at image synthesis tasks. However, these approaches usually employ additional image encoders and rely on the cross-attention mechanism for texture transfer from the garment to the person image, which affects the try-on's efficiency and fidelity. To address these issues, we propose an Texture-Preserving Diffusion (TPD) model for virtual try-on, which enhances the fidelity of the results and introduces no additional image encoders. Accordingly, we make contributions from two aspects. First, we propose to concatenate the masked person and reference garment images along the spatial dimension and utilize the resulting image as the input for the diffusion model's denoising UNet. This enables the original self-attention layers contained in the diffusion model to achieve efficient and accurate texture transfer. Second, we propose a novel diffusion-based method that predicts a precise inpainting mask based on the person and reference garment images, further enhancing the reliability of the try-on results. In addition, we integrate mask prediction and image synthesis into a single compact model. The experimental results show that our approach can be applied to various try-on tasks, e.g., garment-to-person and person-to-person try-ons, and significantly outperforms state-of-the-art methods on popular VITON, VITON-HD databases. Code is available at \href{https://github.com/Gal4way/TPD}{\texttt{https://github.com/Gal4way/TPD}}.
\end{abstract}    
\section{Introduction}
\label{sec:intro}
Image-based virtual try-on has recently attracted significant interest in the research community as online shopping increases in popularity~\cite{han2018viton, wang2018toward, li2021toward, fele2022c, albahar2021pose, huang2022towards, yang2022full, gou2023taming}. The goal of image-based virtual try-on is to replace the clothes in a person image with a specified garment in a photo-realistic manner. It can potentially enhance the customers' online shopping experience significantly; however, this task remains challenging. A key problem is that the reference garment must be naturally deformed to fit the specified person’s body shape and pose. Moreover, the patterns and texture details on the reference garment should be preserved and distorted realistically during the virtual try-on process.

\begin{figure}[!h]
    \centering
    \includegraphics[width=0.47\textwidth]{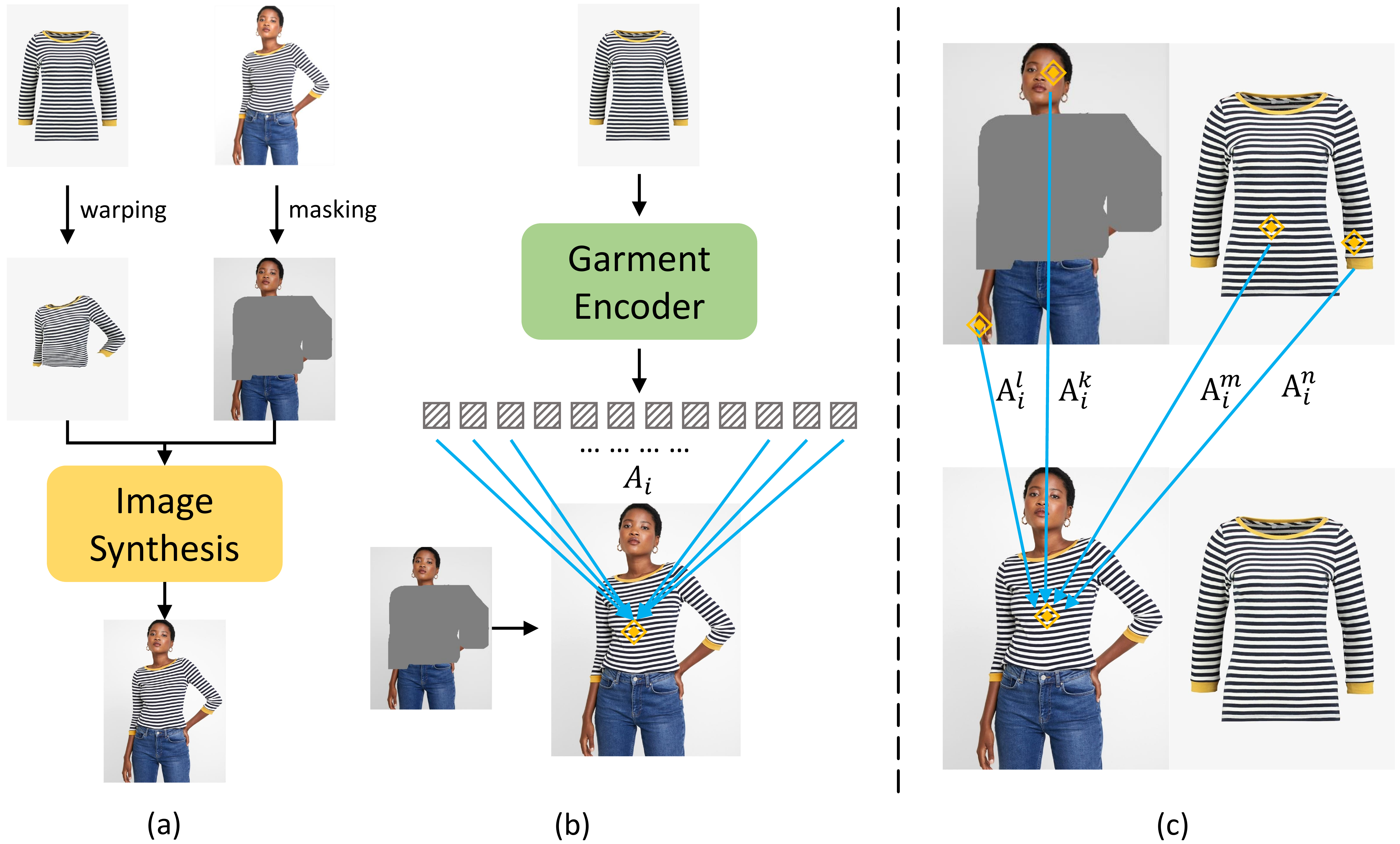}
    \caption{Comparisons between different virtual try-on mechanisms. (a) The warping-based mechanism. (b) The cross-attention-based warping-free mechanism. (c) Our self-attention-based mechanism. $A$ represents the attention weight of a specific query-key pair.}

    \label{fig_comparision_attention}
    \vspace*{-6mm}
\end{figure}

To overcome this challenge, existing methods~\cite{han2018viton, wang2018toward, yu2019vtnfp, choi2021viton, lee2022high, fele2022c, yang2022full, issenhuth2020not, gou2023taming, chen2023size, chopra2021zflow, xie2023gp, yan2023linking} generally perform garment warping before image synthesis, as illustrated in Figure~\ref{fig_comparision_attention}(a). However, garment warping produces artifacts that are difficult to correct in the synthesis stage~\cite{gou2023taming, zhu2023tryondiffusion}. Hence, recent works~\cite{gou2023taming, Li_Wei_Yin_Ma_Kot, zhu2023tryondiffusion, morelli2023ladi} have begun exploring warping-free methods based on the powerful diffusion models~\cite{ho2020denoising, saharia2022photorealistic, rombach2022high}. They typically utilize the cross-attention mechanism~\cite{vaswani2017attention} in the denoising UNet to transfer the textures in the reference garment to the corresponding areas of the person image, as shown in Figure~\ref{fig_comparision_attention}(b). To extract the reference garment's texture features, DCI-VTON~\cite{gou2023taming} and MGD~\cite{baldrati2023multimodal} directly utilize the original CLIP encoder~\cite{radford2021learning}, while LaDI-VTON~\cite{morelli2023ladi} and TryOnDiffusion~\cite{zhu2023tryondiffusion} adopt additional image encoders, e.g., a Vision Transformer (VIT)~\cite{Kolesnikov2021VIT} or an additional UNet~\cite{ronneberger2015u} model.

However, the subject of efficiently generating high-fidelity try-on images remains underexplored. First, extracting features using the CLIP image encoder~\cite{gou2023taming, yang2022paint} results in the loss of fine-grained textures, as this encoder was initially trained to align with the holistic features of coarse captions. In addition, utilizing specialized image encoders~\cite{morelli2023ladi, zhu2023tryondiffusion} increases computational costs. Second, existing methods~\cite{choi2021viton, lee2022high, gou2023taming, zhu2023tryondiffusion, chen2023size, xie2023gp, yan2023linking} generally remove the original garment in the person image through a roughly estimated inpainting mask. While it may not cover every texture in the original person image’s garment, it often removes garment-irrelevant textures, such as tattoos and muscle structures~\cite{zhu2023tryondiffusion}, as shown in the experimentation section. This issue further impacts the try-on results’ fidelity.

Therefore, we propose a Texture-Preserving Diffusion (TPD) model for high-fidelity virtual try-on to address these challenges. First, we propose a Self-Attention-based Texture Transfer (SATT) method. In contrast to existing approaches, we discard garment warping and specialized garment image encoders in our method. Instead, we discover that the original self-attention blocks within the diffusion model are more effective and efficient for garment texture transfer. Specifically, as illustrated in Figure~\ref{fig_comparision_attention}(c), we concatenate the masked person and the reference garment images along the spatial dimension, and the resulting image is fed into the diffusion model. Then, we leverage the powerful self-attention blocks in the Stable Diffusion (SD) model's~\cite{rombach2022high} denoising UNet~\cite{ronneberger2015u} to capture the long-range correlations among pixels in the combined image. This strategy regards the reference garment as the context for the masked person in the same image and enables efficient texture transfer from the garment to the person image in the forward pass of the diffusion model. Moreover, since the UNet contains self-attention blocks with multiple resolutions, it facilitates more effective texture transfer across different feature scales. In the experimentation section, we demonstrate the capability of SATT in generating high-fidelity try-on images with complex textures, patterns, and challenging body pose variations.

Second, we propose a Decoupled Mask Prediction (DMP) method that automatically determines an accurate inpainting area for each person-garment image pair. Since an accurate mask is determined by the original person and the reference garment images, we predict this mask in a decoupled manner. Specifically, DMP iteratively denoises the mask from an initial random noise to an inpainting area determined by the reference garment. We also obtain the area of the original garment in the person image using a human parsing tool. Finally, we use the union of both areas as the final inpainting mask. Unlike existing approaches that adopt mask solely determined by the original person image, the mask predicted by DMP adapts to the garment it encounters, enabling us to preserve as much identity information as possible. In the experimentation section, we demonstrate that DMP preserves fingers, arms and tattoos compared to existing methods, enhancing synthesized images' fidelity.

Our key contributions are summarized as follows. First, we propose a novel diffusion-based and warping-free method that achieves a more efficient and accurate virtual try-on. Second, we explore the coarse inpainting masks' effect on the fidelity of the synthesized images and propose a novel method for accurate mask prediction. Third, our approach consistently outperforms state-of-the-art methods in the realism and coherence of the synthesized images on popular VITON and VITON-HD databases.
\section{Related Work}
\label{sec:relatedworks}
\paragraph{Image-based Virtual Try-On.}
The existing image-based virtual try-on methods can be divided into warping-based and warping-free approaches. 

The warping-based approaches~\cite{han2018viton, wang2018toward, yu2019vtnfp, issenhuth2020not, chopra2021zflow, choi2021viton, lee2022high, fele2022c, yang2022full, gou2023taming, chen2023size, xie2023gp, yan2023linking} perform garment warping before image synthesis. They typically adopt a two-stage framework: the first stage warps the garment image to the body in the person image, while the second synthesizes the final image by fusing the warped garment and the person images. Thin Plate Spline (TPS)~\cite{duchon1977splines, han2018viton, minar2020cp, lee2019viton, yang2020towards}, flow map~\cite{zhou2016view, bai2022single, chopra2021zflow, ge2021parser, han2019clothflow, he2022style}, and landmark~\cite{liu2021toward, xie2020lg, yan2023linking, chen2023size} facilitate garment warping. Regarding the image synthesis stage, one method group promotes the fidelity of synthesized images by providing extra cues like human parsing maps~\cite{yu2019vtnfp,yang2020towards, choi2021viton}, while the other~\cite{choi2021viton,dong2020fashion,fele2022c} improves the image quality by modifying the generative models' structure, like introducing additional normalization layers. Recently, researchers have begun leveraging diffusion models~\cite{rombach2022high} instead of Generative Adversarial Networks (GANs)~\cite{goodfellow2020generative} in the image synthesis stage due to their powerful image generation capabilities~\cite{gou2023taming, Li_Wei_Yin_Ma_Kot}. As a result, they have obtained try-on images of higher quality and realism. The main disadvantage of warping-based methods is the artifacts produced by garment warping, which are difficult to correct in the image synthesis stage.

In contrast, warping-free methods~\cite{baldrati2023multimodal, zhu2023tryondiffusion, morelli2023ladi} are usually diffusion model-based~\cite{ho2020denoising, rombach2022high}. They bypass garment warping to avoid generating artifacts. They typically mask the original garment in the person image and transfer the garment textures to the masked area using an additional image encoder and cross-attention blocks in the diffusion model's denoising UNet. To achieve this goal, Baldrati et al.~\cite{baldrati2023multimodal} adopted the original CLIP text encoder in the SD model to achieve a multi-modal virtual try-on. Similar to Paint-by-Example~\cite{yang2022paint}, Gou et al.~\cite{gou2023taming} replaced the CLIP text encoder with the CLIP image encoder to extract image features as a condition. Additionally, Morelli et al.~\cite{morelli2023ladi} introduced an additional VIT~\cite{Kolesnikov2021VIT} model to supplement the CLIP encoder. However, the CLIP image encoder was pre-trained to align with the holistic features of coarse textual captions; therefore, the extracted features are also coarse and bring in texture loss in the resulting try-on images. Instead of using the off-the-shelf SD model, Zhu et al.~\cite{zhu2023tryondiffusion} trained a new diffusion model from scratch based on their private large-scale database. They also introduced an additional U-Net model to replace the CLIP image encoder that facilitates multi-scale feature extraction from the garment image. However, the enlarged model architecture also incurs additional computational costs.

This paper addresses the fidelity issues in existing warping-free virtual try-on methods. We propose to utilize the original self-attention blocks within the diffusion model to achieve a more powerful and efficient garment texture transfer. We also introduce an approach that automatically determines an accurate inpainting area according to the specific person-garment pair, which enables the model to generate high-fidelity images.

\vspace*{-6mm}
\paragraph{Diffusion Models.} 
Diffusion models~\cite{ho2020denoising, ramesh2022hierarchical, saharia2022photorealistic, rombach2022high} have attracted significant research attention, as they generate high-quality images and enable stable training convergence. The Denoising Diffusion Probabilistic Model (DDPM) was first proposed to model image generation as a diffusion process~\cite{ho2020denoising}. Then, Denoising Diffusion Implicit Models (DDIM)~\cite{song2020denoising} and Pseudo Numerical methods for Diffusion Models (PNDM)~\cite{liu2022pseudo} were proposed to accelerate the generation process by developing new noise schedulers. More recently, latent diffusion models~\cite{rombach2022high} have been introduced to perform the diffusion process in the latent space of a pre-trained Variational Autoencoder (VAE)~\cite{kingma2013auto}, which enables higher computational efficiency and synthesized image quality. 

Latent diffusion models have been applied in various image generation tasks~\cite{zhang2023adding, wu2022tune, gal2022image, karras2023dreampose}, and many studies are aimed at improving the controllability of the generation process. For example, Yang et al.~\cite{yang2022paint} replaced the CLIP text encoder in the SD model with a CLIP image encoder, enabling the model to generate images according to the image condition. Karras et al.~\cite{karras2023dreampose} adopted a pre-trained VAE encoder to supplement the CLIP image encoder, improving the generation of high-fidelity images. Recently, Zhang et al.~\cite{zhang2023adding} proposed the ControlNet model, which introduces an additional network that injects image conditions into the frozen SD model as explicit guidance. ControlNet performs adequately for tasks where the input and output are aligned in the structures, but it may struggle with virtual try-on due to the significant pose differences between the person and garment images.

This study addresses the virtual try-on's challenges based on the SD model. Compared to the above studies, we generate high fidelity try-on images without using specialized image encoders. Moreover, our approach is robust and can manage significant pose differences.

\begin{figure*}
\vspace*{-6mm}
\centering
\includegraphics[width=\textwidth]{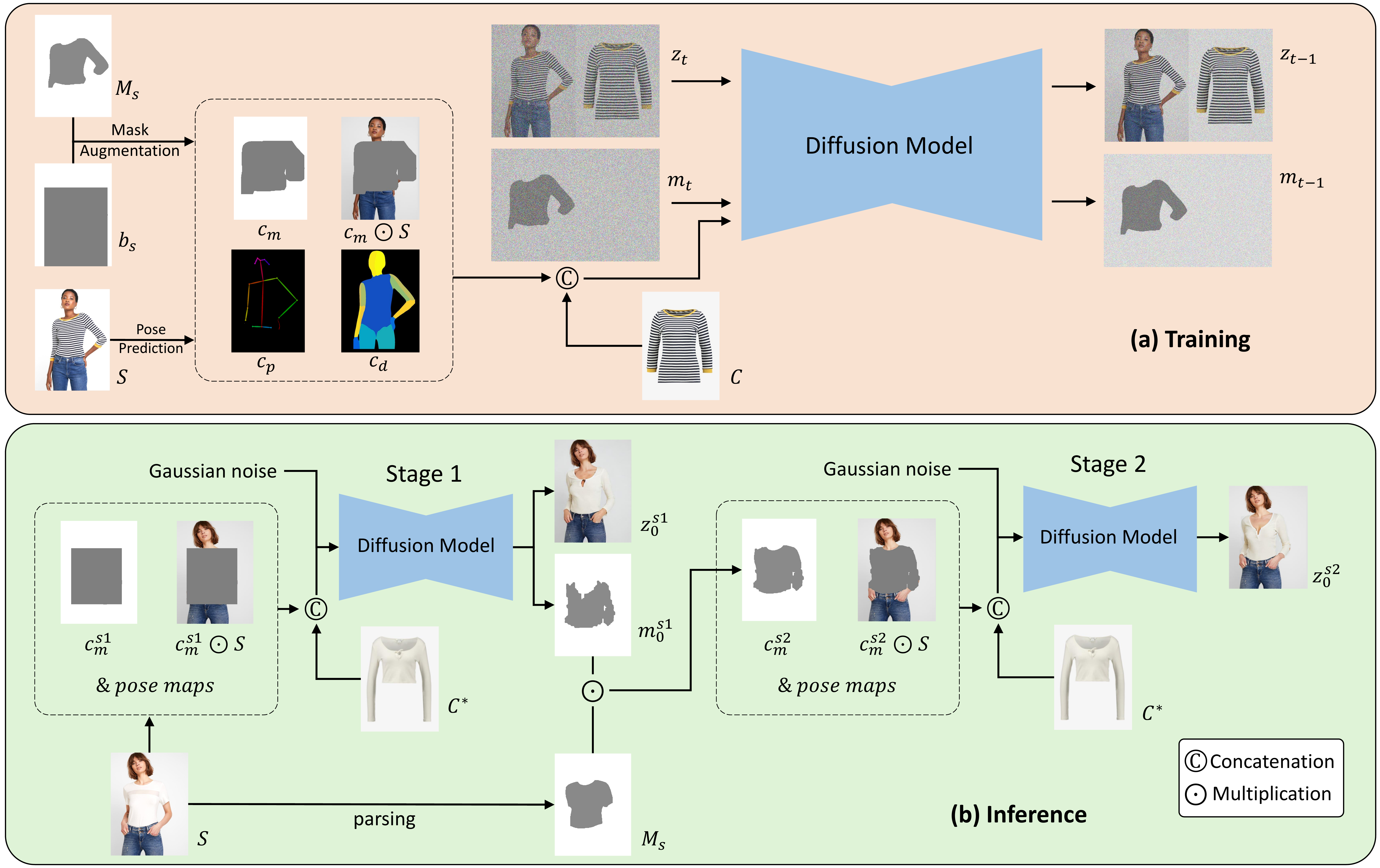}
\caption{An overview of our framework. (a) In the training phase, we begin with the original person image $S$ and a randomly augmented mask $c_m$. $c_m$ is obtained by interpolating between the original clothing area $M_s$ and the bounding box $b_s$. The augmented mask $c_m$, the masked person image $c_m \odot S$, the pose map $c_p$, and the dense pose $c_d$ serve as the auxiliary input for the denoising UNet. Furthermore, the reference garment $C$ is concatenated with each of the auxiliary input along the spatial dimension as the context of the self-attention mechanism. (b) The inference phase is divided into two stages. In the first stage, we predict the clothing area $m_0^{s1}$ for the new garment $C^*$ on the person. We obtain $c_m^{s2}$ via element-wise multiplication between $m_0^{s1}$ and $M_s$. In the second stage, $c_m^{s2}$ is utilized as an accurate inpainting mask, enabling the diffusion model to produce high-fidelity try-on images. For clarity, we omit the predicted concatenated garments from the results of both stages.}
\label{fig_model}
\end{figure*}

\section{Method}
\subsection{Preliminary: Diffusion Models}
DDPMs~\cite{ho2020denoising} iteratively recover images from normally distributed random noise. To improve training and inference speed, recent diffusion models, e.g., SD model~\cite{rombach2022high}, operate in the encoded latent space of a pre-trained autoencoder~\cite{kingma2013auto}. 
SD consists of two core components: a VAE~\cite{kingma2013auto} and a denoising UNet~\cite{ronneberger2015u}. Specifically, the VAE encoder $E$  first encodes the input image $x \in \textbf{R}^{3 \times H \times W}$ into a latent representation $z = E(x) \in \textbf{R}^{4 \times h \times w}$. After $T$ diffusion steps, $z$ generally develops into an isotropic Gaussian noise $z_T$. Then, the text-conditioned denoising UNet $\epsilon_\theta$ is applied to iteratively predict the noise added during each timestep $t=1, ..., T$ and to finally recover the $z'$. The VAE decoder $D$ reconstructs the original image using $z'$ as its input, i.e., $x' = D(z')$. 
For the inpainting task~\cite{yang2022paint}, U-Net uses two more inputs in addition to $z$, i.e., the inpainting mask $m$ and the inpainting background $E((m \odot x)$. 
The objective is defined as follows: 
\begin{equation}  
L_{SD} = \mathbb{E}_{z,\epsilon\sim\mathcal{N}(0,1),t}[\|\epsilon - \epsilon_\theta(z_t, E(m \odot x), m, t, e)\|_2^2],
\label{eq:1}
\end{equation}
where $\epsilon$ represents the ground-truth noise added in this step, $\odot$ denotes the element-wise multiplication, and $e$ signifies the embeddings obtained using a CLIP encoder.

\subsection{Overview}
The overview of our TPD model is presented in Figure~\ref{fig_model}. In this instance, we adopt the SD model~\cite{rombach2022high} as the backbone. We denote the original person image as $S \in \textbf{R}^{3 \times H \times W}$, the reference garment image as $C^* \in \textbf{R}^{3 \times H \times W}$, and the synthesized person image wearing the reference garment as $I^* \in \textbf{R}^{3 \times H \times W}$. In practice, collecting triplet data in the form of $<S, C^*, I^*>$ is challenging. To solve this problem, existing databases~\cite{han2018viton, choi2021viton} usually adopt paired data in the form of $<S,C>$, where $C$ refers to a garment image that contains the same garment worn by the person in $S$, as illustrated in Figure~\ref{fig_model}. 

In the following sections, we introduce the Self-Attention-based Texture Transfer (SATT) method in Section~\ref{SATT} and the Decoupled Mask Prediction (DMP) method in Section~\ref{DMP}, respectively.

\subsection{Self-Attention-based Texture Transfer}
\label{SATT}
The SD model's denoising UNet contains convolutional~\cite{ding2018trunk}, self-attention, and cross-attention blocks in each resolution level. Existing methods typically utilize the cross-attention blocks to achieve garment-to-person texture transfer. Therefore, they focus on promoting the feature extraction power of the specialized garment image encoders~\cite{morelli2023ladi, zhu2023tryondiffusion, baldrati2023multimodal, yang2022paint}, whose outputs serve as the key and value for cross-attention operations. However, enhancing the power of specialized garment image encoders usually incurs additional computational costs~\cite{zhu2023tryondiffusion}. We argue that existing works overlook the potential benefits of the self-attention blocks. 

This section proposes to utilize the inherent self-attention blocks in the denoising UNet for more accurate and efficient virtual try-on. Fundamentally, we regard both the reference garment and the unmasked area in the person image as the context for the inpainting task. Specifically, we first concatenate the garment image $C$ and the masked person image $c_m \odot S$ along the spatial dimension. Then, we feed the resulting image into the UNet. This makes $C$ part of the context in the combined image. Accordingly, the task of the diffusion model becomes reconstructing both the person and garment images from random Gaussian noise, as illustrated in Figure~\ref{fig_model}. As a result, the UNet’s convolutional blocks extract the garment’s textures, and the self-attention blocks efficiently transfer textures from the garment to the person image. As illustrated in Figure~\ref{fig_comparision_attention}, the self-attention operation can be represented as follows:
\begin{equation}
\text{Attention}(Q, K, V) = \text{softmax}\left(\frac{QK^T}{\sqrt{d}}\right)V,
\end{equation}
where $Q,K,V \in \mathbb{R}^{p \times d}$ are stacked vectors reshaped from the same latent feature map, $p$ is the number of pixels in the feature map, and $d$ represents the vector dimension. In this way, the correlation between each pixel pair in the feature map is considered, naturally achieving texture transfer from the garment area to the person area within the same image.

Alternatively, $C$ and $c_m \odot S$ can be concatenated along the channel dimension. However, as mentioned in~\cite{zhu2023tryondiffusion}, the pixels in $C$ and $c_m\odot S$ are not spatially aligned; therefore, the textures in $C$ can hardly be transferred to the masked area in $c_m\odot S$ using convolution or self-attention operations. Section~\ref{sec:experiments} demonstrates that our strategy performs significantly better than the concatenation operation along the channel dimension.

\subsection{Decoupled Mask Prediction}
\label{DMP}
Existing methods~\cite{gou2023taming, choi2021viton, wang2018toward, han2018viton, yu2019vtnfp} generally employ a mask to remove the original garment in the person image. Therefore, the accuracy of this mask is vital to the virtual try-on task's performance. However, existing methods tend to roughly estimate one mask for each person image and apply it to all reference garments~\cite{choi2021viton}. As illustrated in Figure~\ref{fig_comparison_mask}, this rough mask may cover some background and body-part areas, resulting in unnecessary loss of information. These issues affect the fidelity of the synthesized try-on image $I^*$.

We propose a method to predict an accurate mask for each specific $<S,C^*>$ pair to solve this problem. Assuming that the person is simultaneously wearing the original and new garments, the accurate inpainting mask is equal to the union of both clothing areas. Since the original clothing area $M_s$ can be obtained from $S$ using human parsing, our approach aims to predict the new garment's clothing area. 

In addition to predicting the latent $z$ for the image synthesis task, our method incorporates an additional channel $m$ dedicated to predicting the clothing area of the new garment on the target person, as illustrated in Figure~\ref{fig_model}. Notably, the training data is in the form of $<S,C>$, and the predicted mask in the training phase is precisely the clothing area of the original garment in $S$. In comparison, the data in the inference phase is in the form of $<S,C^*>$. Therefore, we adopt the following two-stage prediction pipeline during testing. 
As illustrated in Figure~\ref{fig_model}, in the first stage, we utilize a bounding box as the initial inpainting mask $c_m^{s1}$. Our model iteratively predicts a coarse try-on image and the clothing area $m_0^{s1}$ for the new garment $C^*$ iteratively from random Gaussian noise.
In the second stage, we utilize the union of $m_0^{s1}$ and $M_s$, resulting in an accurate inpainting mask $c_m^{s2}$ for the current person-garment image pair. This accurate mask enables us to preserve the pixels in the background and body-part areas irrelevant to the new garment. Our model produces high-fidelity images with this mask, as shown in the third and last columns in Figure~\ref{fig_comparison_mask}.

Moreover, we introduce the following two strategies to enhance our model's robustness. First, we adopt the pose map $c_p$~\cite{8765346} and dense pose $c_d$~\cite{guler2018densepose} of $S$ as auxiliary input along with $c_m$ and $c_m \odot S$. $c_p$ and $c_d$ provide the body pose and shape information in the masked area. Each of them is also concatenated with the reference garment image along the spatial dimension. Second, we augment the initial mask in the training phase by randomly interpolating between $M_s$ and the bounding box $b_s$. This is because our model encounters coarse and accurate masks in the first and second inference stages, respectively. This augmentation strategy makes our model robust and enables it to tackle the varied shapes of inpainting masks observed in the testing phase.

In summary, we obtain accurate inpainting masks via DMP, allowing us to achieve warping-free virtual try-on with minimal modification to the original person image.

\begin{table}
\centering
    \caption{The quantitative comparisons between our method and state-of-the-art methods on VITON~\cite{han2018viton} and VITON-HD~\cite{choi2021viton} databases.}
\label{tab_quantitative_VITONandVITON-HD}
\resizebox{0.5\textwidth}{!}{
    \begin{tabular}{llcccccc}
    \toprule
    {Database} & {Method} & {SSIM$\uparrow$} & {FID$\downarrow$} &{LPIPS$\downarrow$} \\
    \midrule
    {} & CP-VTON~\cite{wang2018toward} &  0.78 & 24.43 & - \\
    {} & ClothFlow~\cite{han2019clothflow} &  0.84 & 14.43 & - \\
    {} & ACGPN~\cite{yang2020towards}  &  0.84 & 15.67  & 0.11 \\
    {VITON} & SDAFN~\cite{bai2022single}  &  0.85 & 10.55  & 0.09 \\
    {} & PF-AFN~\cite{ge2021parser}  &  0.87 & 10.09 & 0.08  \\
    {} & Paint-by-Example~\cite{yang2022paint}  &  0.83 & 12.56 & 0.12  \\
    % \cmidrule[0.5pt](rl){1-1}
    % \cmidrule[0.5pt](rl){2-4}
    % \cmidrule[0.5pt](rl){5-7}
    % \midrule
    {} & \textbf{Ours} & \textbf{0.89} & \textbf{9.58} & \textbf{0.07}\\
    \midrule  
    {} & CP-VTON~\cite{wang2018toward} &  0.79 & 30.25 & 0.14  \\
    {} & PF-AFN~\cite{ge2021parser}  &  0.85 & 11.30 & 0.08  \\
    {} & VITON-HD~\cite{choi2021viton}  &  0.84 & 11.65 & 0.11  \\
    {VITON-HD} & HR-VITON~\cite{lee2022high}  &  0.87 & 10.91 & 0.10  \\
    {} & LaDI-VTON~\cite{morelli2023ladi}  &  0.87 & 9.41 & 0.09  \\ 
    {} & DCI-VTON~\cite{gou2023taming}  &  0.88 & 8.78 & 0.08  \\ 
    {} & Paint-by-Example~\cite{yang2022paint}  &  0.84 & 12.15 & 0.13  \\ 
    {} & \textbf{Ours} & \textbf{0.90} & \textbf{8.54} & \textbf{0.07} \\
    \bottomrule
    \end{tabular}
}
\vspace*{-4mm}
\end{table}

\section{Experiments}
\label{sec:experiments}
\begin{figure}[!h]
    \centering
    \includegraphics[width=0.47\textwidth]{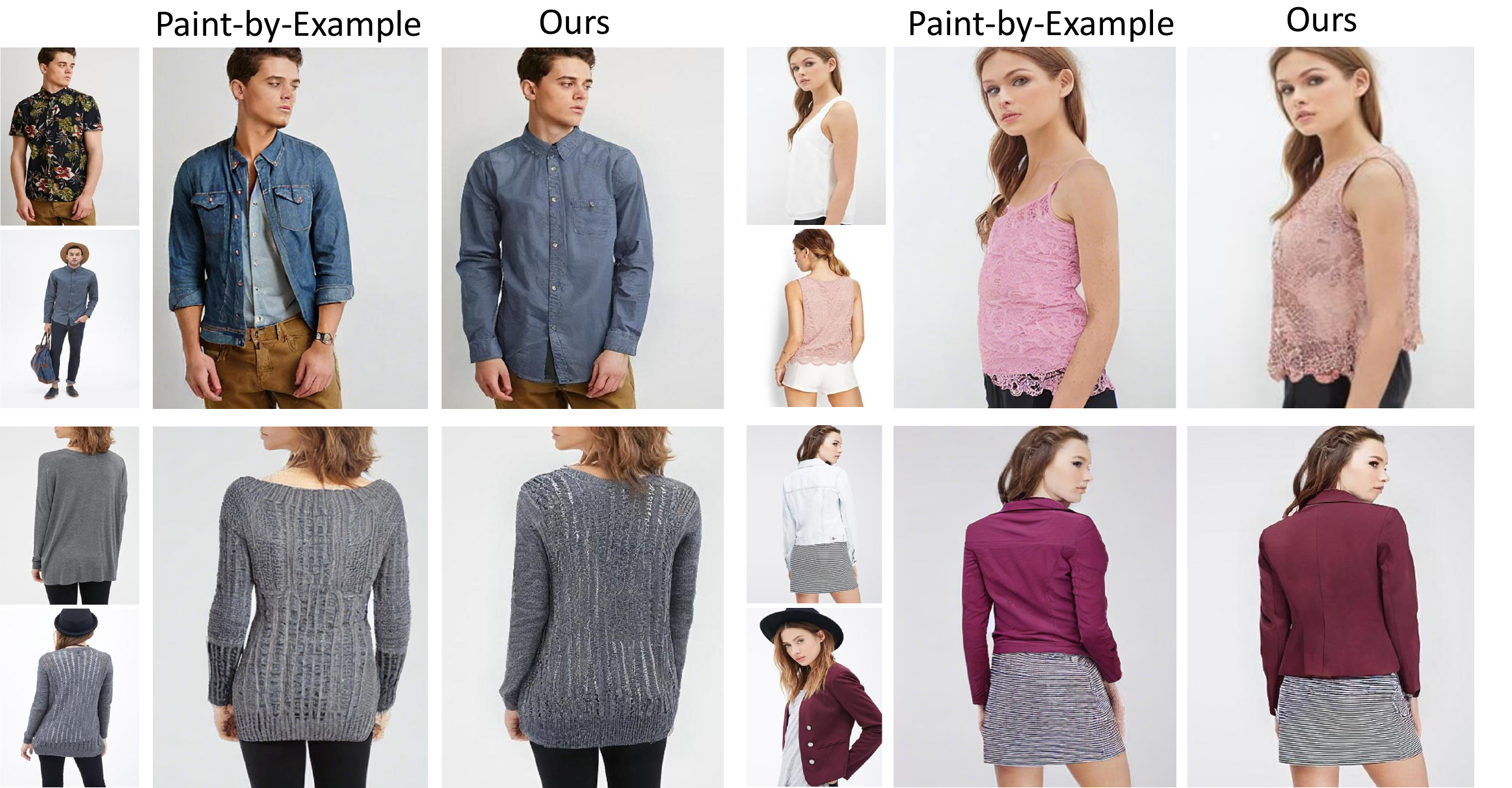}
    \caption{The qualitative comparisons between our method and Paint-by-Example~\cite{yang2022paint} on DeepFashion~\cite{liuLQWTcvpr16DeepFashion} database.} 
    \label{quality_DeepFashion}
    \vspace*{-4mm}
\end{figure}

\paragraph{Databases and Metrics.} Experiments are conducted on three virtual try-on benchmarks: VITON~\cite{han2018viton}, VITON-HD~\cite{choi2021viton}, and DeepFashion~\cite{Liu_Luo_Qiu_Wang_Tang_2016}. VITON contains training and testing sets of 14,221 and 2,032 image pairs, respectively. Each image pair has a front-view photo of a female and a reference garment. The image resolution is 256 $\times$ 192 pixels. VITON-HD is similar to VITON except that its image resolution is 1024 $\times$ 768 pixels. In our experiments, we resize all images to 512 $\times$ 384 pixels for comparison. Moreover, we conduct additional experiments on DeepFashion for the person-to-person virtual try-on task, which involves fitting the garment on a person to another person's body. This is significantly more challenging and the experimental results are illustrated in Figure~\ref{quality_DeepFashion} and Figure~\ref{fig_comparison_mask}. 

We compare our model's performance with state-of-the-art methods in paired and unpaired settings. In the paired setting, the person in $S$ wears the same garment as the reference image. In the unpaired setting, the reference garment is different from the original one in $S$. Structural Similarity (SSIM)~\cite{wang2004image}, Learned Perceptual Image Patch Similarity (LPIPS)~\cite{johnson2016perceptual}, and Frechet Inception Distance (FID)~\cite{heusel2017gans} are utilized to measure the accuracy and realism of the synthesized images. Similar to existing studies~\cite{yang2020towards, bai2022single, ge2021parser, choi2021viton, lee2022high}, the SSIM score and LPIPS are used for the paired setting, while the FID score is used for the unpaired. 

\paragraph{Implementation Details.}
Similar to existing virtual try-on studies~\cite{choi2021viton, morelli2023ladi}, we employ OpenPose~\cite{8765346}, Graphonomy~\cite{Gong2019Graphonomy}, and Detectron2~\cite{wu2019detectron2} to extract the pose map, human-parsing maps, and dense pose of the person, respectively. We train our model with the Adam optimizer~\cite{Kingma_Ba_2014}, and the learning rate is set to 1e-5.

\begin{table}
\centering
    \caption{The ablation study on each key TPD component on VITON-HD~\cite{choi2021viton} database.}
\label{tab_ablation}
\resizebox{0.5\textwidth}{!}{
    \begin{tabular}{llcccccc}
    \toprule
    {Method} & {SSIM$\uparrow$} & {FID$\downarrow$} &{LPIPS$\downarrow$} \\
    \midrule
    w/o SATT &  0.85 & 11.34 & 0.12 \\
    Channel-dim Transfer &  0.85 & 10.95 & 0.11 \\
    w/o DMP  &  0.88 & 9.08  & 0.08 \\
    w/o Mask Augmentation  &  0.80 & 27.24  & 0.19 \\
    \midrule
    \textbf{Ours}  &  \textbf{0.90} & \textbf{8.54}  & \textbf{0.07} \\
    \bottomrule
    \end{tabular}
}
\vspace*{-4mm}
\end{table}

\begin{figure*}[!h]
    \centering
    \includegraphics[width=\textwidth]{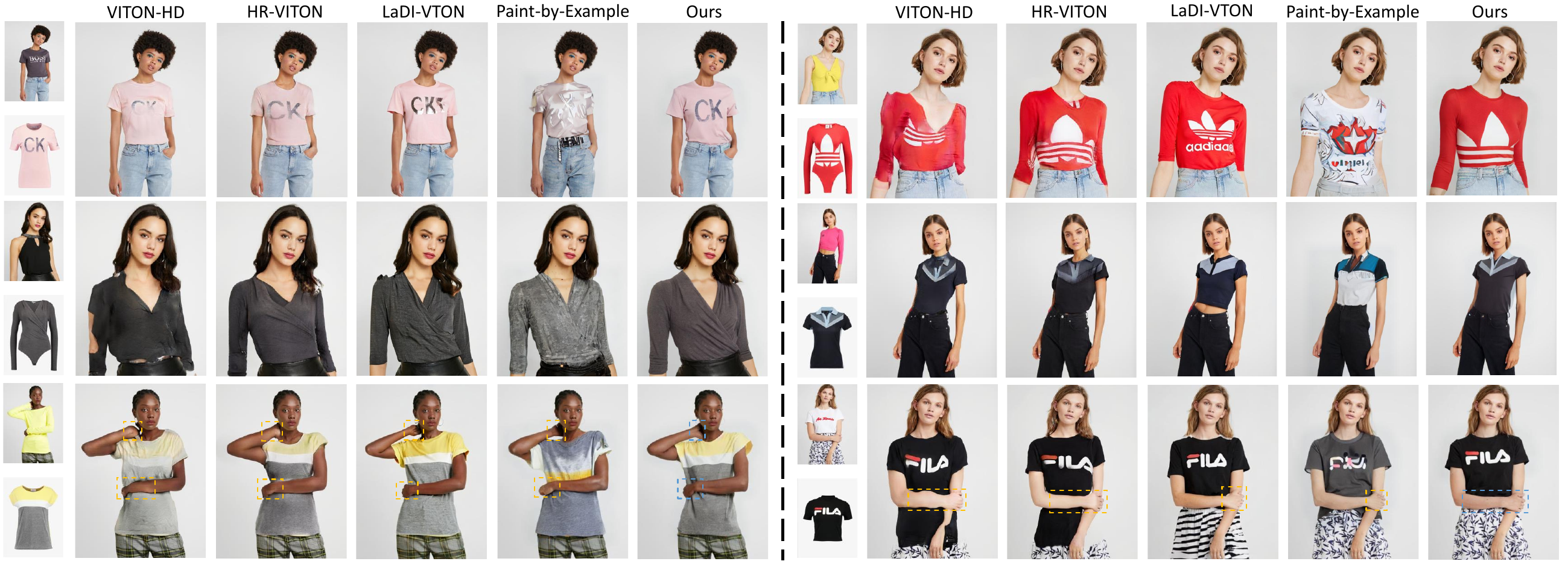}
    \caption{The qualitative comparisons between our method and state-of-the-art methods on VITON-HD~\cite{choi2021viton} database.}
    \label{fig_qualitative_VITON-HD}
\end{figure*}

\begin{figure*}[!h]
    \centering
    \includegraphics[width=\textwidth]{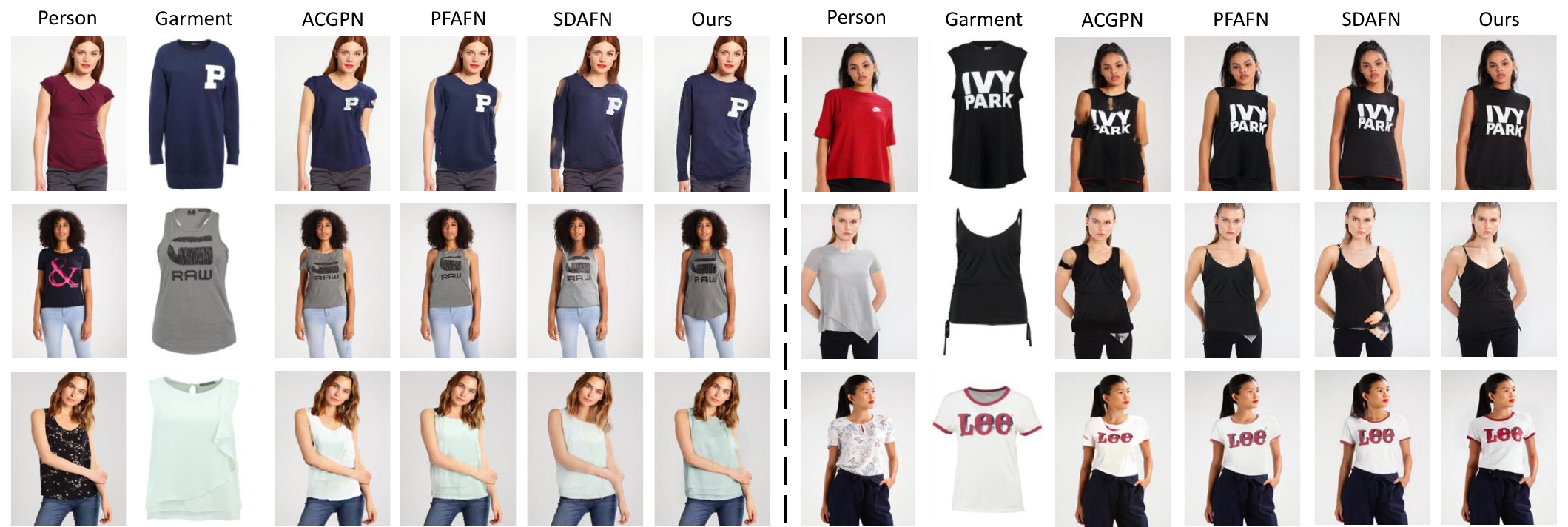}
    \caption{The qualitative comparisons between our method and state-of-the-art methods on VITON~\cite{han2018viton} database.}
    \label{fig_qualitative_VITON}
\end{figure*}

\subsection{Qualitative Comparisons}
Figure~\ref{fig_qualitative_VITON-HD} and Figure~\ref{fig_qualitative_VITON} depict the qualitative comparisons between TPD and state-of-the-art methods including ACGPN~\cite{yang2020towards}, PF-AFN~\cite{ge2021parser}, SDAFN~\cite{bai2022single}, VITON-HD~\cite{choi2021viton}, HR-VITON~\cite{lee2022high}, LaDI-VTON~\cite{morelli2023ladi}, and the diffusion-based inpainting method Paint-by-Example~\cite{yang2022paint}. Try-On Diffusion~\cite{zhu2023tryondiffusion} is excluded from the comparisons as it is not open sourced and it was trained on a large-scale private database in a person-to-person try-on setting. This makes fair comparisons with this method infeasible.

We observe that TPD generates higher quality and fidelity images than other methods. First, existing methods tend to produce artifacts for garments with complex textures, e.g., texts and logos in the first row of Figure~\ref{fig_qualitative_VITON-HD}. There are two main reasons for these artifacts: (1) the warping operations tend to generate artifacts, and (2) the image encoders used by these methods lose the fine-grained textures in the reference garment. Second, the performance of existing methods decreases for person images with challenging poses. As illustrated in the third row of Figure~\ref{fig_qualitative_VITON-HD}, these methods generate distorted fingers or arms because they adopt inaccurate masks when removing the original garment in the person image, resulting in the loss of human body part information.

In comparison, TPD can generate high-quality try-on images with fewer artifacts. One main reason is that our self-attention-based texture transfer method is warping-free and enables efficient multi-scale feature extraction from the garment image. Another reason is that we utilize DMP to determine the precise inpainting area based on the person image and the reference garment, as illustrated in Figure~\ref{fig_comparison_mask}. This enables us to modify the original person image within minimal pixels, leading to high-fidelity results for images with challenging poses.

\subsection{Quantitative Results}
Table~\ref{tab_quantitative_VITONandVITON-HD} presents the quantitative comparisons between TPD and state-of-the-art methods, including CP-VTON~\cite{wang2018toward}, ClothFlow~\cite{han2019clothflow}, ACGPN~\cite{yang2020towards}, SDAFN~\cite{bai2022single}, PF-AFN~\cite{ge2021parser}, VITON-HD~\cite{choi2021viton}, HR-VITON~\cite{lee2022high}, Dci-VTON~\cite{gou2023taming}, LaDI-VTON~\cite{morelli2023ladi}, and the diffusion-based inpainting method Paint-by-Example~\cite{yang2022paint}. This table shows that TPD consistently achieves the best performance on both VITON~\cite{han2018viton} and VITON-HD~\cite{choi2021viton} databases. Specifically, it achieves the leading FID scores, demonstrating that the images it generates are of higher-quality. Moreover, it achieves the best SSIM and LPIPS scores, indicating that it generates try-on images with the correct semantics.

\subsection{Ablation Study}
\begin{figure}[!h]
    \centering
    \includegraphics[width=0.47\textwidth]{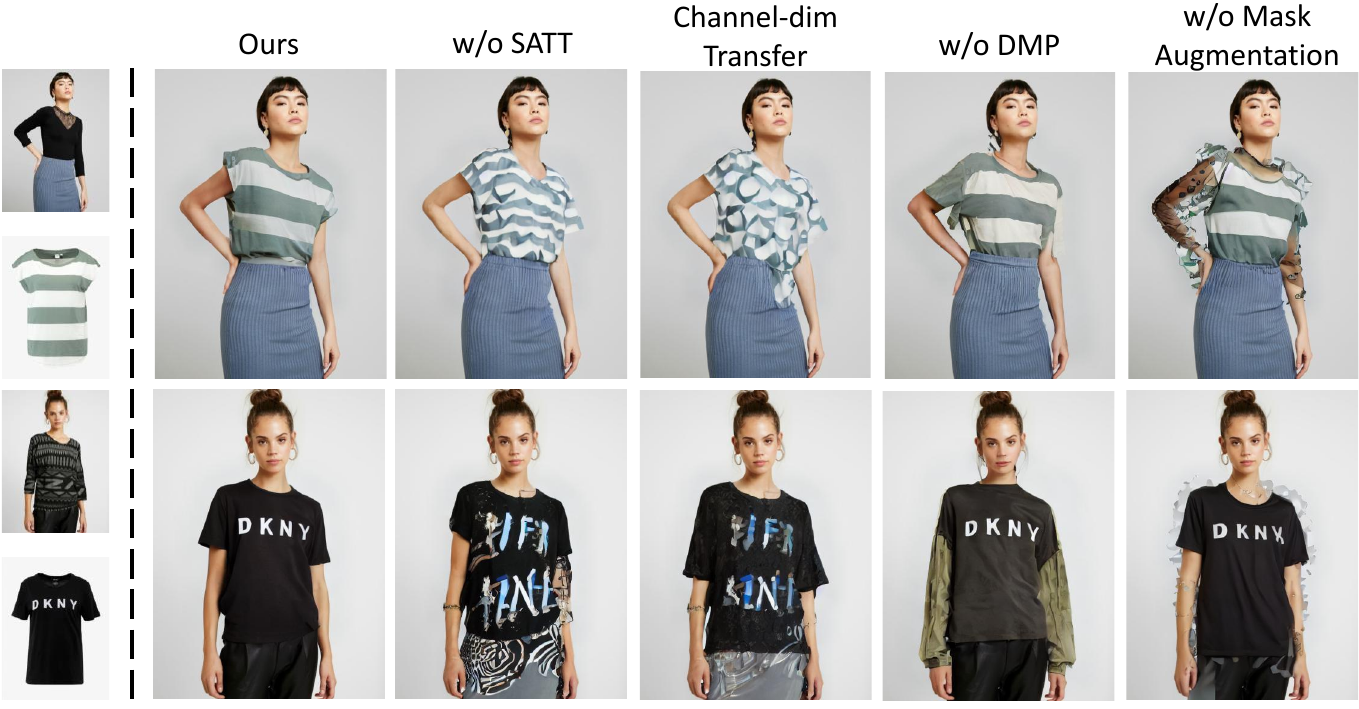}
    \caption{The ablation study on each TPD key component on VITON-HD~\cite{choi2021viton} database.}
    \label{fig_ablation}
    \vspace*{-2mm}
\end{figure}

\begin{figure}[!h]
    \centering
    \includegraphics[width=0.48\textwidth]{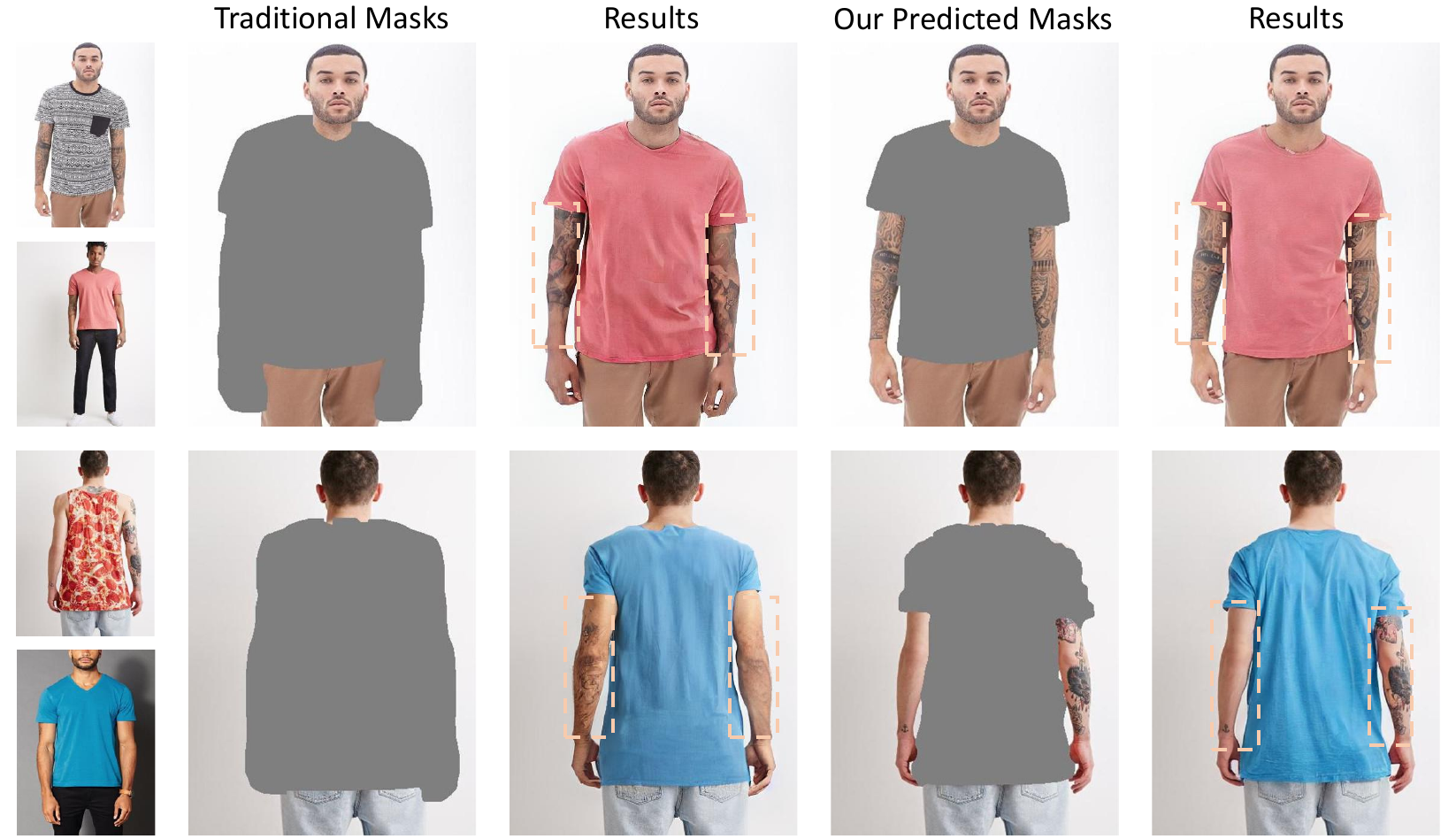}
    \caption{The comparisons between synthesized try-on images with the traditional masks and our predicted masks on DeepFashion~\cite{liuLQWTcvpr16DeepFashion} database.} 
    \label{fig_comparison_mask}
    \vspace*{-6mm}
\end{figure}

We perform an ablation study in Table~\ref{tab_ablation}, Figure~\ref{fig_ablation} and Figure~\ref{fig_comparison_mask} to justify each key TPD component's effectiveness. 

First, we validate SATT's effectiveness. Instead of using SATT, we extract garment features via the CLIP image encoder, as introduced in Section~\ref{sec:relatedworks}. Accordingly, texture transfer is accomplished using the cross-attention blocks in the denoising UNet. This method is denoted as `w/o SATT' in Table~\ref{tab_ablation} and Figure~\ref{fig_ablation}. We demonstrate that the performance of this method is notably poorer than that of SATT. This is because it is difficult for the CLIP image encoder to extract fine-grained texture features from the reference garment image, as this encoder was pre-trained to align with the holistic features of coarse captions~\cite{radford2021learning}.

Second, we further validate the importance of concatenating the garment and masked person images along the spatial dimension to SATT. Specifically, we adopt the alternative strategy mentioned in Section~\ref{SATT}, which concatenates the two images along the channel dimension. This method is denoted as `Channel-dim Transfer' in Table~\ref{tab_ablation} and Figure~\ref{fig_ablation}. Both qualitative and quantitative results show that SATT leads to results of higher-fidelity. This is because the pixels in the garment and masked person images are not spatially aligned, which makes texture transfer across channels difficult. In contrast, spatial concatenation makes the garment a part of the context in the masked person image, enabling easier and more accurate texture transfer to the masked area.

Third, we demonstrate DMP's effectiveness. Specifically, we remove DMP from the TPD framework and use traditional masks instead~\cite{choi2021viton}. This method is named `w/o DMP' in Table~\ref{tab_ablation} and Figure~\ref{fig_ablation}. Figure~\ref{fig_ablation} and Figure~\ref{fig_comparison_mask} show that compared with traditional masks, those predicted by DMP enable us to obtain improved try-on results, including preserving body details, e.g., arms or tattoos. This is because DMP predicts accurate masks, which minimizes the loss of irrelevant textures to the try-on task in the synthesized image results. 

Finally, we verify the effectiveness of DMP's mask augmentation strategy. This experiment is represented as `w/o Mask Augmentation' in Table~\ref{tab_ablation} and Figure~\ref{fig_ablation}. It is shown that the model produces notable artifacts in the try-on results without the mask augmentation strategy. This is because the model only encounters coarse masks during the training stage. Hence, it cannot handle the accurate masks viewed in the second stage during inference. As illustrated in the second column of Figure~\ref{fig_ablation}, mask augmentation effectively removes these artifacts in the synthesized images.

\section{Conclusion and Limitations}
In this paper, we propose a Texture-Preserving Diffusion (TPD) model for high-fidelity virtual try-on without using specialized garment image encoders. Our approach concatenates the person and reference garment images along the spatial dimension and uses the combined image as the input for the Stable Diffusion model's denoising UNet. This enables accurate feature transfer from the garment to the person image using the inherent self-attention blocks in the diffusion model. To preserve the background and human body-part details as much as possible, our model also predicts a precise inpainting mask based on the reference garment and the original person images, further enhancing the fidelity of the synthesized results. Furthermore, TPD can be widely applied to garment-to-person and person-to-person virtual try-on tasks. The extensive experiments show that our approach achieves state-of-the-art performance on VITON~\cite{han2018viton} and VITON-HD~\cite{choi2021viton} databases. This work also has certain limitations. For example, images in nearly all databases for this task have single-color background. Therefore, our model’s performance on images with more complex backgrounds is to be explored in the future. Details can be found in the supplementary materials.

\paragraph{Broader Impacts}
Virtual try-on methods can generate try-on images based on the person and reference garment images, which means it is significant for real-world applications like online shopping and e-commerce. Moreover, our approach may be applied to other diffusion model-based image editing tasks, such as image inpainting, image-to-image translation. This adaptability broadens its utility to the community, paving the way for more advanced image synthesis and editing works. To the best of our knowledge, this work does not have obvious negative social impacts. 

\paragraph{Acknowledgement}
This work was supported by the National Natural Science Foundation of China under Grant 62076101, Guangdong Basic and Applied Basic Research Foundation under Grant 2023A1515010007, the Guangdong Provincial Key Laboratory of Human Digital Twin under Grant 2022B1212010004, the TCL Young Scholars Program.

% WARNING: do not forget to delete the supplementary pages from your submission 
% \input{sec/X_suppl}


\begin{thebibliography}{9}
{
    \small
    
    \bibitem{fele2022c}
    B. Fele, A. Lampe, P. Peer, V. Struc. C-vton: Context-driven image-based virtual try-on network. In {\it{WACV}}, 2022.
    
    \bibitem{han2019clothflow}
    X. Han, X. Hu, W. Huang, M. Scott. Clothflow: A flow-based model for clothed person generation. In {\it{ICCV}}, 2019.
    
    \bibitem{yang2020towards}
    H. Yang, R. Zhang, X. Guo, W. Liu, W. Zuo, P. Luo. Towards photo-realistic virtual try-on by adaptively generating-preserving image content. In {\it{CVPR}}, 2020.
    

    \bibitem{vaswani2017attention}
    A. Vaswani, N. Shazeer, N. Parmar, J. Uszkoreit, L. Jones, A. Gomez. Attention is all you need. In {\it{NeurIPS}}, 2017.
    
    \bibitem{liuLQWTcvpr16DeepFashion}
    Z. Liu, P. Luo, S. Qiu, X. Wang, X. Tang. DeepFashion: Powering Robust Clothes Recognition and Retrieval with Rich Annotations. In {\it{CVPR}}, 2016.
    

    \bibitem{liu2021toward}
    G. Liu, D. Song, R. Tong, M. Tang. Toward realistic virtual try-on through landmark guided shape matching. In {\it{AAAI}}, 2021.
    
    \bibitem{xie2020lg}
    Z. Xie, J. Lai, X. Xie. LG-VTON: Fashion landmark meets image-based virtual try-on. In {\it{PRCV}}, 2020.
    

    \bibitem{lee2022high}
    S. Lee, G. Gu, S. Park, S. Choi, J. Choo. High-Resolution Virtual Try-On with Misalignment and Occlusion-Handled Conditions. In {\it{ECCV}}, 2022.
    


    \bibitem{bai2022single}
    S. Bai, H. Zhou, Z. Li, C. Zhou, H. Yang. Single stage virtual try-on via deformable attention flows. In {\it{ECCV}}, 2022.
    
    \bibitem{he2022style}
    S. He, Y. Song, T. Xiang. Style-based global appearance flow for virtual try-on. In {\it{CVPR}}, 2022.
    
    \bibitem{li2021toward}
    K. Li, M. Chong, J. Zhang, J. Liu. Toward accurate and realistic outfits visualization with attention to details. In {\it{CVPR}}, 2021.
    
    \bibitem{wang2018toward}
    B. Wang, H. Zheng, X. Liang, Y. Chen, L. Lin, M. Yang. Toward characteristic-preserving image-based virtual try-on network. In {\it{ECCV}}, 2018.


    
    \bibitem{choi2021viton}
    S. Choi, S. Park, M. Lee, J. Choo. Viton-hd: High-resolution virtual try-on via misalignment-aware normalization. In {\it{CVPR}}, 2021.
    
  
    \bibitem{han2018viton}
    X. Han, Z. Wu, Z. Wu, R. Yu, L. Davis. Viton: An image-based virtual try-on network. In {\it{CVPR}}, 2018.
    
    \bibitem{song2020denoising}
    J. Song, C. Meng, S. Ermon. Denoising diffusion implicit models. \href{https://arxiv.org/abs/2010.02502}{arXiv:2010.02502}, 2020.
    
    
    \bibitem{issenhuth2020not}
    T. Issenhuth, J. Mary, C. Calauzenes. Do not mask what you do not need to mask: a parser-free virtual try-on. In {\it{ECCV}}, 2020.
    

    \bibitem{ho2020denoising}
    J. Ho, A. Jain, P. Abbeel. Denoising diffusion probabilistic models. In {\it{NeurIPS}}, 2020.
    
    \bibitem{yang2022full}
    H. Yang, X. Yu, Z. Liu. Full-range virtual try-on with recurrent tri-level transform. In {\it{CVPR}}, 2022.
    
    \bibitem{liu2022pseudo}
    L. Liu, Y. Ren, Z. Lin, Z. Zhao. Pseudo numerical methods for diffusion models on manifolds. \href{https://arxiv.org/abs/2202.09778}{arXiv:2202.09778}, 2022.
    
    \bibitem{rombach2022high}
    R. Rombach, A. Blattmann, D. Lorenz, P. Esser. High-resolution image synthesis with latent diffusion models. In {\it{CVPR}}, 2022.
    

    \bibitem{yang2022paint}
    B. Yang, S. Gu, B. Zhang, T. Zhang, X. Chen, X. Sun, D. Chen, F. Wen. Paint by Example: Exemplar-based Image Editing with Diffusion Models. \href{https://arxiv.org/abs/2211.13227}{arXiv:2211.13227}, 2022.
    

    
    \bibitem{yu2019vtnfp}
    R. Yu, X. Wang, X. Xie. Vtnfp: An image-based virtual try-on network with body and clothing feature preservation. In {\it{ICCV}}, 2019.
    
    \bibitem{chopra2021zflow}
    A. Chopra, R. Jain, M. Hemani, B. Krishnamurthy. Zflow: Gated appearance flow-based virtual try-on with 3d priors. In {\it{ICCV}}, 2021.
    
    \bibitem{ge2021parser}
    Y. Ge, Y. Song, R. Zhang, C. Ge, W. Liu, P. Luo. Parser-free virtual try-on via distilling appearance flows. In {\it{CVPR}}, 2021.
    
    \bibitem{goodfellow2020generative}
    I. Goodfellow, J. Pouget-Abadie, M. Mirza, B. Xu, D. Warde-Farley, S. Ozair, A. Courville, Y. Bengio. Generative adversarial networks. In {\it{Communications of the ACM}}, 2020.
    
    \bibitem{gal2022image}
    R. Gal, Y. Alaluf, Y. Atzmon, O. Patashnik, A. Bermano, G. Chechik, D. Cohen-Or. An image is worth one word: Personalizing text-to-image generation using textual inversion. \href{https://arxiv.org/abs/2208.01618}{arXiv:2208.01618}, 2022.
    

    \bibitem{Kingma_Ba_2014}
    D. Kingma, J. Ba. Adam: A Method for Stochastic Optimization. \href{https://arxiv.org/abs/1412.6980}{arXiv:1412.6980}, 2022.


    \bibitem{guler2018densepose}
    R. Güler, N. Neverova, I. Kokkinos. Densepose: Dense human pose estimation in the wild. In {\it{CVPR}}, 2018.


    \bibitem{8765346}
    Z. Cao, T. Simon, S. Wei, Y. Sheikh. OpenPose: Realtime Multi-Person 2D Pose Estimation using Part Affinity Fields. In {\it{TPAMI}}, 2019.

    \bibitem{ding2018trunk}
    C. Ding, D. Tao. Trunk-Branch Ensemble Convolutional Neural Networks for Video-Based Face Recognition. In {\it{TPAMI}}, 2018.

    \bibitem{minar2020cp}
    M. Minar, T. Tuan, H. Ahn, P. Rosin, Y. Lai. Cp-vton+: Clothing shape and texture preserving image-based virtual try-on. In {\it{CVPR Workshops}}, 2020.
    
    \bibitem{radford2021learning}
    A. Radford, J. Kim, C. Hallacy, A. Ramesh, G. Goh, S. Agarwal, G. Sastry, A. Askell, P. Mishkin, J. Clark. Learning transferable visual models from natural language supervision. In {\it{ICML}}, 2021.
    
    \bibitem{zhang2023adding}
    L. Zhang, M. Agrawala. Adding conditional control to text-to-image diffusion models. \href{https://arxiv.org/abs/2302.05543}{arXiv:2302.05543}, 2023.


    
    \bibitem{johnson2016perceptual}
    J. Johnson, A. Alahi, L. Fei-Fei. Perceptual losses for real-time style transfer and super-resolution. In {\it{ECCV}}, 2016.
    
    \bibitem{duchon1977splines}
    J. Duchon. Splines minimizing rotation-invariant semi-norms in Sobolev spaces. In {\it{CTFSV}}, 1977.
    
    \bibitem{lee2019viton}
    H. Lee, R. Lee, M. Kang, M. Cho, G. Park. LA-VITON: A network for looking-attractive virtual try-on. In {\it{ICCV}}, 2019.
    
    \bibitem{zhou2016view}
    T. Zhou, S. Tulsiani, W. Sun, J. Malik, A. Efros. View synthesis by appearance flow. In {\it{ECCV}}, 2016.
    
    \bibitem{ramesh2022hierarchical}
    A. Ramesh, P. Dhariwal, A. Nichol, C. Chu, M. Chen. Hierarchical text-conditional image generation with clip latents. \href{https://arxiv.org/abs/2204.06125}{arXiv:2204.06125}, 2022.
    
    \bibitem{saharia2022photorealistic}
    C. Saharia, W. Chan, S. Saxena, L. Li, J. Whang, E. Denton, K. Ghasemipour, R. Gontijo Lopes, B. Karagol Ayan, T. Salimans. Photorealistic text-to-image diffusion models with deep language understanding. In {\it{NeurIPS}}, 2022.
    

    
    \bibitem{wu2022tune}
    J. Wu, Y. Ge, X. Wang, W. Lei, Y. Gu, W. Hsu, Y. Shan, X. Qie, M. Shou. Tune-A-Video: One-Shot Tuning of Image Diffusion Models for Text-to-Video Generation. \href{https://arxiv.org/abs/2212.11565}{arXiv:2212.11565}, 2022.
    

    \bibitem{karras2023dreampose}
    J. Karras, A. Holynski, T. Wang, I. Kemelmacher-Shlizerman. DreamPose: Fashion Image-to-Video Synthesis via Stable Diffusion. \href{https://arxiv.org/abs/2304.06025}{arXiv:2304.06025}, 2023.


    \bibitem{kingma2013auto}
    D. Kingma, M. Welling. Auto-encoding variational bayes. \href{https://arxiv.org/abs/1312.6114}{arXiv:1312.6114}, 2013.
    
    \bibitem{ronneberger2015u}
    O. Ronneberger, P. Fischer, T. Brox. U-net: Convolutional networks for biomedical image segmentation. In {\it{MICCAI}}, 2015.
    

    
    \bibitem{wang2004image}
    Z. Wang, A. Bovik, H. Sheikh, E. Simoncelli. Image quality assessment: from error visibility to structural similarity. {\it{TIP, 13}}, 2004.
    
    \bibitem{heusel2017gans}
    M. Heusel, H. Ramsauer, T. Unterthiner, B. Nessler, S. Hochreiter. Gans trained by a two time-scale update rule converge to a local nash equilibrium. In {\it{NeurIPS}}, 2017.
    
    
    \bibitem{gou2023taming}
    J. Gou, S. Sun, J. Zhang, J. Si, C. Qian, L. Zhang. Taming the Power of Diffusion Models for High-Quality Virtual Try-On with Appearance Flow. \href{https://arxiv.org/abs/2308.06101}{arXiv:2308.06101}, 2023.

    \bibitem{baldrati2023multimodal}
    A. Baldrati, D. Morelli, G. Cartella, M. Cornia, M. Bertini, R. Cucchiara. Multimodal Garment Designer: Human-Centric Latent Diffusion Models for Fashion Image Editing. In {\it{ICCV}}, 2023.

    \bibitem{dong2020fashion}
    H. Dong, X. Liang, Y. Zhang, X. Zhang, X. Shen, Z. Xie, B. Wu, J. Yin. Fashion editing with adversarial parsing learning. In {\it{CVPR}}, 2020.
    

    \bibitem{Kolesnikov2021VIT}
    A. Dosovitskiy, L. Beyer, A. Kolesnikov, D. Weissenborn, X. Zhai, T. Unterthiner, M. Dehghani, M. Minderer, G. Heigold, S. Gelly. An image is worth 16x16 words: Transformers for image recognition at scale. \href{https://arxiv.org/abs/2010.11929}{arXiv:2010.11929}, 2020.

    \bibitem{wu2019detectron2}
    Y. Wu, A. Kirillov, F. Massa, W. Lo, R. Girshick. Detectron2. \url{https://github.com/facebookresearch/detectron2}, 2019.

    
    \bibitem{Liu_Luo_Qiu_Wang_Tang_2016}
    Z. Liu, P. Luo, S. Qiu, X. Wang, X. Tang. Deepfashion: Powering robust clothes recognition and retrieval with rich annotations. In {\it{CVPR}}, 2016.


    \bibitem{Gong2019Graphonomy}
    K. Gong, Y. Gao, X. Liang, X. Shen, M. Wang, L. Lin. Graphonomy: Universal Human Parsing via Graph Transfer Learning. In {\it{CVPR}}, 2019.


    
    \bibitem{chen2023size}
    C. Chen, Y. Chen, H. Shuai, W. Cheng. Size Does Matter: Size-aware Virtual Try-on via Clothing-oriented Transformation Try-on Network. In {\it{ICCV}}, 2023.
    
    \bibitem{Li_Wei_Yin_Ma_Kot}
    Z. Li, P. Wei, X. Yin, Z. Ma, A. Kot. Virtual Try-On with Pose-Garment Keypoints Guided Inpainting. In {\it{ICCV}}, 2023.
    
    \bibitem{xie2023gp}
    Z. Xie, Z. Huang, X. Dong, F. Zhao, H. Dong, X. Zhang, F. Zhu, X. Liang. GP-VTON: Towards General Purpose Virtual Try-on via Collaborative Local-Flow Global-Parsing Learning. In {\it{CVPR}}, 2023.
    
    \bibitem{yan2023linking}
    K. Yan, T. Gao, H. Zhang, C. Xie. Linking Garment With Person via Semantically Associated Landmarks for Virtual Try-On. In {\it{CVPR}}, 2023.
    
    \bibitem{morelli2023ladi}
    D. Morelli, A. Baldrati, G. Cartella, M. Cornia, M. Bertini, R. Cucchiara. LaDI-VTON: Latent Diffusion Textual-Inversion Enhanced Virtual Try-On. \href{https://arxiv.org/abs/2305.13501}{arXiv:2305.13501}, 2023.
    
    \bibitem{zhu2023tryondiffusion}
    L. Zhu, D. Yang, T. Zhu, F. Reda, W. Chan, C. Saharia, M. Norouzi, I. Kemelmacher-Shlizerman. TryOnDiffusion: A Tale of Two UNets. In {\it{CVPR}}, 2023.


    \bibitem{albahar2021pose}
    B. Albahar, J. Lu, J. Yang, Z. Shu, E. Shechtman, J. Huang. Pose with Style: Detail-preserving pose-guided image synthesis with conditional stylegan. In {\it{TOG}}, 2021.

    \bibitem{huang2022towards}
    Z. Huang, H. Li, Z. Xie, M. Kampffmeyer, X. Liang. Towards hard-pose virtual try-on via 3d-aware global correspondence learning. In {\it{ANIPS}}, 2022.




%     \bibliographystyle{ieeenat_fullname}
%     \bibliography{main}
}
\end{thebibliography}
\end{document}